\documentclass[10pt,twocolumn,letterpaper]{article}

\usepackage[pagenumbers]{cvpr} 

\usepackage{natbib}
\setcitestyle{numbers,compress}
\usepackage{times}
\usepackage{epsfig}
\usepackage{graphicx}
\usepackage{amsmath}
\usepackage{amssymb}
\usepackage{booktabs}
\usepackage{multirow}
\usepackage{comment}
\usepackage{xcolor}
\usepackage{comment}
\definecolor{cvprblue}{rgb}{0.21,0.49,0.74}
\usepackage[pagebackref,breaklinks,colorlinks,citecolor=cvprblue]{hyperref}
\usepackage{graphicx}
\usepackage{longtable}
\usepackage[accsupp]{axessibility} 

\newcommand{\denselist}{\itemsep 0pt\parsep=0pt\partopsep 0pt}
\newcommand{\mypara}[1]{\noindent\textbf{#1}}

\definecolor{brightred}{RGB}{255, 80, 106}
\definecolor{jointred}{RGB}{192, 0, 0}
\definecolor{jointblue}{RGB}{68, 114, 196}
\definecolor{jointyellow}{RGB}{255, 192, 0}
\definecolor{jointgrey}{RGB}{118, 113, 113}
\definecolor{nodepink}{RGB}{236, 181, 212}
\definecolor{nodeyellow}{RGB}{255, 230, 153}

\def\paperID{3769}
\def\confName{CVPR}
\def\confYear{2024}
\title{CAGE: Controllable Articulation GEneration}

\author{
Jiayi Liu \quad Hou In Ivan Tam \quad Ali Mahdavi-Amiri \quad Manolis Savva \\
Simon Fraser University\\
\small\href{https://3dlg-hcvc.github.io/cage/}{3dlg-hcvc.github.io/cage}
}

\begin{document}

\newcommand{\figfirstpagefigure}{
\vspace{-3em}
\begin{center}
\captionsetup{type=figure}
\includegraphics[width=\textwidth]{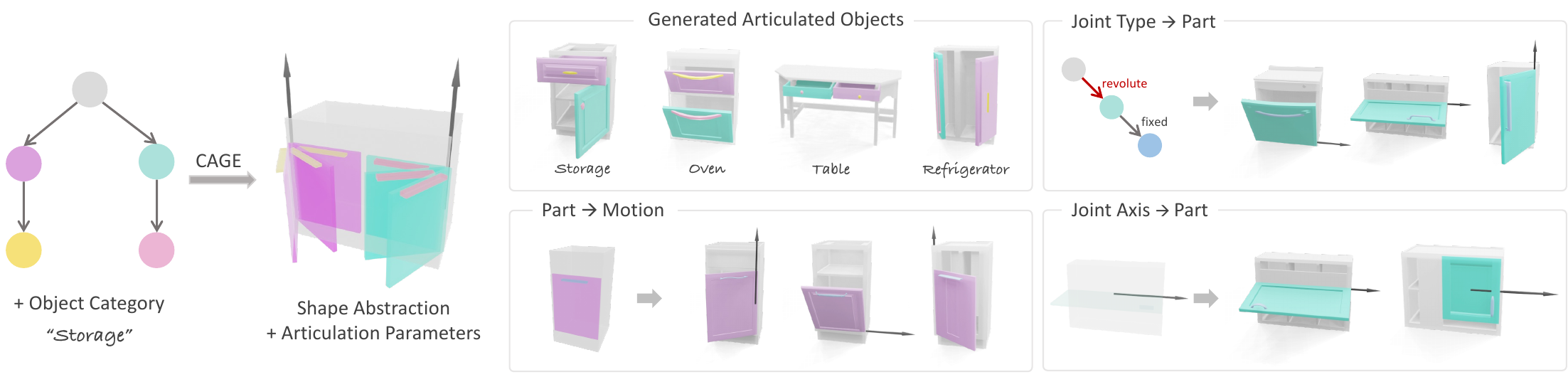}
\captionof{figure}{
We present CAGE: a user-controllable generative model for 3D articulated objects.
\textbf{Left}: given an object category label and a directed graph describing the interconnections among constituent parts, our model generates an abstraction of the articulated object specifying both geometry and motion parameters for each part.
\textbf{Right}: the generated shape abstraction combined with appropriately constrained part retrieval allows for generating high-quality articulated objects under various user-specified constraints.
Users can specify a desired object category, part shape, articulation type, or articulation axis and obtain generated objects that respect the provided constraints.
}
\label{fig:teaser}
\end{center}
\vspace{0.5em}
}
\twocolumn[{
\maketitle
\figfirstpagefigure
}]


\begin{abstract}

We address the challenge of generating 3D articulated objects in a controllable fashion.
Currently, modeling articulated 3D objects is either achieved through laborious manual authoring, or using methods from prior work that are hard to scale and control directly.
We leverage the interplay between part shape, connectivity, and motion using a denoising diffusion-based method with attention modules designed to extract correlations between part attributes.
Our method takes an object category label and a part connectivity graph as input and generates an object's geometry and motion parameters.
The generated objects conform to user-specified constraints on the object category, part shape, and part articulation.
Our experiments show that our method outperforms the state-of-the-art in articulated object generation, producing more realistic objects while conforming better to user constraints.
\end{abstract}    
\section{Introduction}
\label{sec:intro}

Articulated objects are ubiquitous in real-world scenes.
Kitchen cabinetry, refrigerators, storage drawers, and wardrobes are a few examples.
Thus, digital 3D models of these objects are useful for a variety of tasks in robotics~\cite{fu2022robotube,yu2023gamma}, 3D vision~\cite{wald2020beyond,jiang2022opd,mao2022multiscan}, and embodied AI systems~\cite{xiang2020sapien,szot2021habitat}.
Unfortunately, despite this clear value in many research areas, 3D models of articulated objects are predominantly authored manually and existing datasets of such models are few and relatively small~\cite{mo2019partnet,liu2022akb}.

Unsurprisingly, there is much recent work on reconstructing articulated objects from real-world observations~\cite{jiang2022ditto,hsu_dittohouse,tseng2022cla-nerf,jiayi2023paris} and on predicting how object parts can articulate for existing 3D object models~\cite{li2019category,yan_rpmnet,wang2018shape2motion}.
However, these methods either rely on laborious acquisition of real-world data, or assume the existence of large datasets of 3D objects with sufficiently complete part geometry to enable articulation.
These assumptions limit the scalability and practical utility of these approaches.

An alternative strategy recently proposed in NAP~\cite{lei2023nap} is to learn a generative model for articulated 3D objects that can directly generate a complete articulated object.
Although the NAP generative model conceptually supports conditional generation of articulated objects for partially specified inputs, as we will see it often fails to respect input constraints, and exhibits limited controllability.
This limitation stands in stark contrast with the natural desire for user-driven control over the output of generative models, especially for highly structured and compositional output as is the case with articulated objects.

In this paper, we tackle the challenge of controllable articulated object generation.
To address the limitations of prior work and enable fine-grained control for 3D articulated object generation, we take a directed graph and category label describing a desired object as input conditions.
Since articulated objects are composed of hierarchies of rigidly moving parts organized in a kinematic chain it is natural and straightforward for designers to use an abstraction such as a graph when creating a new articulated 3D object.
We then develop a denoising diffusion probabilistic model (DDPM)~\cite{ho2020ddpm} based generative model for 3D articulated objects that:
1) disentangles graph structure and part attributes by representing parts as a set of bounding primitives with motion parameters associated with the input graph;
2) models a joint distribution of articulation and shape abstraction among parts; and
3) controls the generation using the input graph structure and object category, and additional conditions on desired part attributes.

One of our key insights is that lifting part geometry to a high-level shape abstraction is effective in succinctly capturing important shape-motion correlations.
Our experiments show that this abstraction coupled with a series of appropriately designed attention modules improve joint modeling of parts and motion compared to prior work which attempts to capture detailed part geometry~\cite{lei2023nap}.
Another insight is that the position of key ``actionable'' parts (e.g., a door handle that is grasped to open the door) in relation to other parts provides a strong signal about likely motion patterns.
For example, should a handle be positioned on the top-left corner of a door, the door is likely to rotate around an axis either on the right or bottom sides.
We leverage this insight by incorporating actionable parts into the \textit{articulation} graph to form an augmented \textit{action} graph, with each actionable part connected to and influencing a parent part.

Our quantitative and qualitative evaluations compare our proposed method against ablations and baselines from prior work, and show that our method generates more realistic and more complex articulated 3D objects, exhibiting physically plausible variations.
Our method also demonstrates better compatibility with various conditional input scenarios enabling better user-controlled generation. In summary: 
\begin{itemize}
    \item We present a generative model for articulated objects that learns a joint distribution over part shape and motion, under the constraints of graph structure and object category.
    \item We design a denoising network that enables strong conditioning through a series of attribute-attribute level attentions to inject user constraints effectively.
    \item Our evaluation with several proposed metrics shows that our method generates samples of higher quality and achieves better controllability compared to prior work.
\end{itemize}

\section{Related Work}
\label{sec:relwork}

\begin{figure*}
    \includegraphics[width=\textwidth]{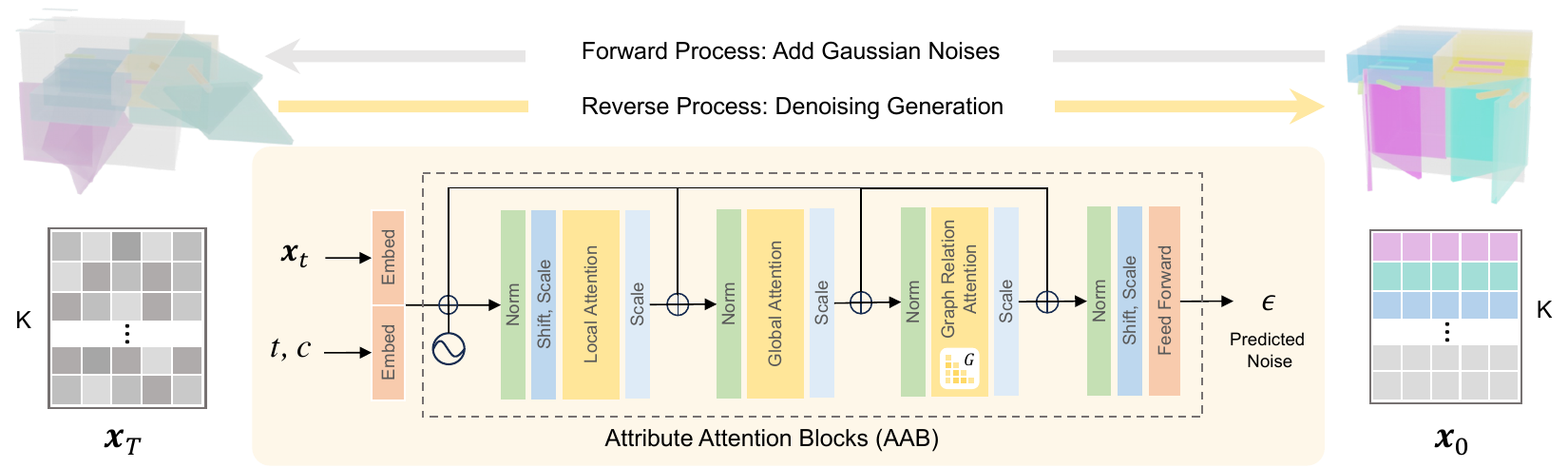}
    \vspace{-5mm}
    \caption{Method overview. Our generative model is based on DDPM~\cite{ho2020ddpm}. In the forward pass, Gaussian noise is iteratively added to corrupt the data from $\mathbf{x}_0$ to random noise $\mathbf{x}_T$. During the reverse process, our denoiser (in yellow highlight) predicts the residual noise to be subtracted from the input data $\mathbf{x}_t$ at timestep $t$ conditioned on the category label $c$ and a graph adjacency $G$ as an attention mask injected in the Graph Relation Attention module. All the timesteps share the same denoiser that is built on layers of our Attribute Attention Blocks.
    }
    \label{fig:overview}
\end{figure*}

\mypara{Generation of structured objects.}
There is prior work on structure-aware generative models for 3D objects that models part geometry.
\citet{G2L18} use a GAN to generate semantically labeled voxel object parts and refine the shape with an auto-encoder.
Similarly, \citet{SAGnet19} use a VAE to encode part geometry and pairwise part relations into a latent code and decode it to generate objects.
These methods do not explicitly consider the part structure hierarchy.
\citet{grass17} represent the part hierarchies as binary trees of symmetry hierarchy~\cite{symhier11} and train a generative recursive autoencoder to encode parts and their geometry.
\citet{structurenet19} improve upon this work by removing the binary tree constraint and represent objects in n-ary graphs for robust generation of more complex objects.
SDM-NET~\cite{sdmnet19} synthesizes deformable meshes enabling more detailed object part generation.
DSG-Net~\cite{dsgnet20} disentangles part geometry and structure for fine-grained controllable generation.

Our work also addresses the challenge of modeling geometric relations between structured components in a controllable generative model.
However, we also generate articulation parameters and thus tackle the additional challenge of generating physically plausible part motions.

\mypara{Conditional diffusion for structured geometry.}
Diffusion models lack effective control over outputs from noise alone.
Therefore, various techniques condition diffusion models on inputs such as text, sketches, low-resolution images, etc.
SDEdit~\cite{meng2022sdedit} injects a guide image into the noise, while \citet{Voynov2023sketch} use a sketch as a condition. 
ControlNet~\cite{zhang2023adding} takes inputs such as edge or depth maps and adjusts Stable Diffusion~\cite{rombach2022high} weights without significant perturbation.
Text-conditioned image and 3D shape generation is currently the most popular.
Various methods~\cite{nichol2021glide,rombach2022high,ramesh2022hierarchical,saharia2022photorealistic} leverage classifier-free guidance~\cite{ho2022classifier}, text-image embeddings such as CLIP~\cite{radford2021learning}, or BERT~\cite{devlin2018bert}.
DreamFusion~\cite{poole2022dreamfusion} uses text-to-image diffusion and image-based guidance to generate 3D shape neural fields~\cite{mildenhall2021nerf}.
Magic3D~\cite{lin2023magic3d} extends this approach to 3D meshes using neural marching tetrahedra~\cite{shen2021deep}.
Voxels~\cite{sella2023vox,li2023diffusion} and point clouds~\cite{nichol2022point} are also employed for 3D object generation based on text prompts.

Unlike images and unstructured shapes, generation of structured 2D or 3D geometry such as articulated objects requires \emph{part-to-part} relations to be captured in a representation such as a graph.
Such relations are not modeled in image or shape representations such as NeRF, voxel grids, point clouds, or polygonal meshes.
A common approach to overcome this challenge uses transformer-based diffusion and incorporates conditioning through attention mechanisms~\cite{peebles2023dit,sha2023house,tang2023diffuscene}.
Our approach is inspired by this line of work, and focuses on the design of a multi-stage attention mechanism enabling fine-grained attention over part-to-part attributes and conditioning through user-provided constraints on the parts and their attributes.
\mypara{Articulated 3D object modeling.}
\citet{jiang2022ditto} introduced Ditto to build digital twins of articulated objects from point clouds.
\citet{carto23} and \citet{rearticulable23} reconstruct articulated objects from a single stereo RGB observation and point cloud videos, respectively.
PARIS~\cite{jiayi2023paris} simultaneously reconstructs part geometry and articulation parameters from multi-view images.
However, these reconstruction-based methods require real-life observations, limiting their scalability and practicality.

Recently, \citet{lei2023nap} proposed NAP to tackle the task of 3D articulated object generation using a part relation tree formalism.
This is similar to our approach, however NAP requires postprocessing to obtain a valid articulation tree and performs quite poorly in conditional generation, often disregarding input constraints.
Moreover, NAP’s fully connected graph has difficulty handling objects with a large number of parts, which limits its adaptability.
Our work addresses these shortcomings and focuses specifically on a variety of conditional generation scenarios, which are particularly important in practical use.
\section{Method}
\label{sec:methods}

Given an object category label $c$ and a graph structure $G$ represented as an adjacency matrix, 
our objective is to generate an object within category $c$ that is comprised of $N$ parts that adhere to the graph.
We learn the joint distribution of shape abstraction and articulation with a diffusion model using a series of attention modules to capture the interrelation among shape-motion attributes effectively. 
\Cref{fig:overview} provides an overview of our generation pipeline and denoising network architecture.


\subsection{Preliminaries}
\label{subsec: preliminaries}

Diffusion models work by corrupting training data via successive addition of Gaussian noise in a forward process, and then learn to recover the original data via iterative denoising in a reverse process. 
Our work follows the original DDPM ~\cite{ho2020ddpm} formulation.
In the forward pass, given a real articulated object $\mathbf{x}_0$ from an underlying distribution $q(\mathbf{x}_0)$, a series of noisy samples $\mathbf{x}_t$ is obtained by gradually adding Gaussian noise $\epsilon\sim\mathcal{N}(0, \mathbf{I})$:
\begin{equation}
    q(\mathbf{x}_t|\mathbf{x}_{t-1}) := \mathcal{N}(\mathbf{x}_t;\sqrt{\alpha_t}\mathbf{x}_{t-1}, (1-\alpha_t)\mathbf{I}),
\end{equation}
where $t=1,\dots,N$ indicates the denoising step and $\alpha_t$ is determined by a noise variance scheduler. 
Practically, the noisy sample $\mathbf{x}_t$ is obtained by $\mathbf{x}_t=\sqrt{\bar{\alpha_t}}\mathbf{x}_0+\sqrt{1-\bar{\alpha_t}}\boldsymbol{\epsilon}_t$, where $\bar{\alpha_t}=\prod_{s=1}^t\alpha_s$.
In the reverse process, the denoising model aims to predict the noise added in each step $\boldsymbol{\epsilon}_\theta(\mathbf{x}_t, t)$.

\mypara{Training loss.}
The training objective of a diffusion model is to minimize the negative log-likelihood of the data by maximizing the variational lower bound.
Following DDPM~\cite{ho2020ddpm}, a simplified training objective is used as our training loss:
\begin{equation}
    \mathcal{L} = \mathbb{E}_{t, \mathbf{x}_0, \boldsymbol{\epsilon}_t}\left[\left\|\boldsymbol{\epsilon}_t-\boldsymbol{\epsilon}_\theta(\sqrt{\bar{\alpha_t}}\mathbf{x}_0+\sqrt{1-\bar{\alpha_t}}\boldsymbol{\epsilon}_t, t)\right\|^2\right]
\end{equation}
Building on this, we train our model to be conditioned on the graph structure $G$ and object category label $c$.

\subsection{Data Parameterization}
\label{subsec:param}

The design of the data representation is crucial when working with diffusion models.
Given a graph structure as a constraint for generation, our denoising target $\mathbf{x}_T$ is represented as a set of part parameters associated with the input graph (see \Cref{fig:attention} top right).
Each part is represented as a node with 5 node attributes.
The attributes are aligned to be an $M$-dimensional vector by repeating if necessary.
The node attributes include:
\begin{itemize}
    \item \textbf{Part bounding box}: we canonicalize all objects to a ``resting'' state (i.e. doors and drawers closed).
    The canonicalization simplifies the problem such that each part bounding box can be assumed to be axis-aligned.
    We represent each axis-aligned bounding box with the 3D positions of the box maximum and minimum corners.
    \item \textbf{Joint type}: we categorize joints into 5 types as \textit{fixed}, \textit{revolute}, \textit{prismatic}, \textit{continuous}, or \textit{screw} joint. 
    The \textit{fixed} joint is for non-rigid parts, \textit{continuous} is a rotation-only joint without limits, while \textit{revolute} is bounded at two ends. 
    \textit{Screw} joints exhibit both unbounded rotational motion and translational motion.
    We represent the joint type as a scalar which we expand to an $M$ dimensional vector.
    \item \textbf{Joint axis}: each articulation is constrained by an axis. We represent each axis direction with a 3D unit vector and the position of the axis origin with a 3D vector.
    \item \textbf{Joint range}: we represent joint range with a 2D vector (left and right bounds), associated with the joint type.
    For continuous and revolute joints this indicates the rotation angle limits and for prismatic and screw joints this indicates the translation distance limits.
    \item \textbf{Semantic label}: Each part is associated with one of 8 semantic category labels (i.e., base, drawer, door, tray, shelf, knob, wheel, and handle), represented by a scalar which we expand to an $M$ dimensional vector.
\end{itemize}
Formally, let $P=\{p_1, p_2, ..., p_N\}$ denote an articulated object with $N$ parts to be generated. 
Each part $p_i$ is represented as $p_i = \{a_{i,j}|a_{i, j}\in\mathbb{R}^M, 1\leq j\leq5\}$, where $a_{i,j}$ denotes the $j^{th}$ attribute for part $i$. 
To generate objects with parts of variable length and leverage the diffusion model, we pad the nodes to a maximum number $K$. 
In summary, the denoising target is a vector of node attributes $\textbf{x}=\{a_{i,j}\}\in\mathbb{R}^{5\times K\times M}$. 
We define $K$ as $32$ and $M$ as $6$. 

\subsection{Denoising Network}
\label{subsec: denoisenet}

Our denoising network uses a series of Attribute Attention Blocks (AAB).
The components in each AAB are shown in \Cref{fig:overview}. 
We design three attention modules interleaved with adaptive layer normalization to inject object category and graph adjacency as constraints in the conditional generation process.
We discuss each component below.

\mypara{Feature embedding.}
Each AAB takes a vector of node attributes $\mathbf{x}_t$ as input, with the timestep $t$, object category label $c$, and graph adjacency $G$ as conditions.
Each node attribute serves as a token in the attention modules after embedding.
We embed the $j^{th}$ attributes for part $i$ ($a_{i,j}$) to a feature vector $\hat{a}_{i,j}\in\mathbb{R}^{128}$ along with two kinds of positional encoding: 
1) indicates the attribute type, ranging from $1$ to $5$; 
2) indicates the node it belongs to, ranging from $1$ to $K$.
The timestep $t$ and category label $c$ are embedded as a $128D$ feature vector through a linear layer.

\mypara{Norm layer.} 
Adaptive layer norm (adaLN)~\cite{xu2019undln} has been adopted in GANs~\cite{brock2019largescale, karras2019style} and diffusion models with a U-Net denoiser~\cite{dhar2021diffbeatgan}. 
Recent conditional image generation work~\cite{peebles2023dit} showed that the adaLN-Zero variant improves condition injection by initializing each residual block as the identity function.
Specifically, in addition to regressing scale and shift parameters, dimension-wise scaling parameters are regressed and applied immediately prior to residual connections.
We follow this design in our attribute attention blocks. 
Once the timestep and category label are embedded in the feature vector as $\hat{t}$ and $\hat{c}$, we pass them into an additional embedding layer to learn all the scaling and shift parameters used between attention layers.

\begin{figure}
    \includegraphics[width=\linewidth]{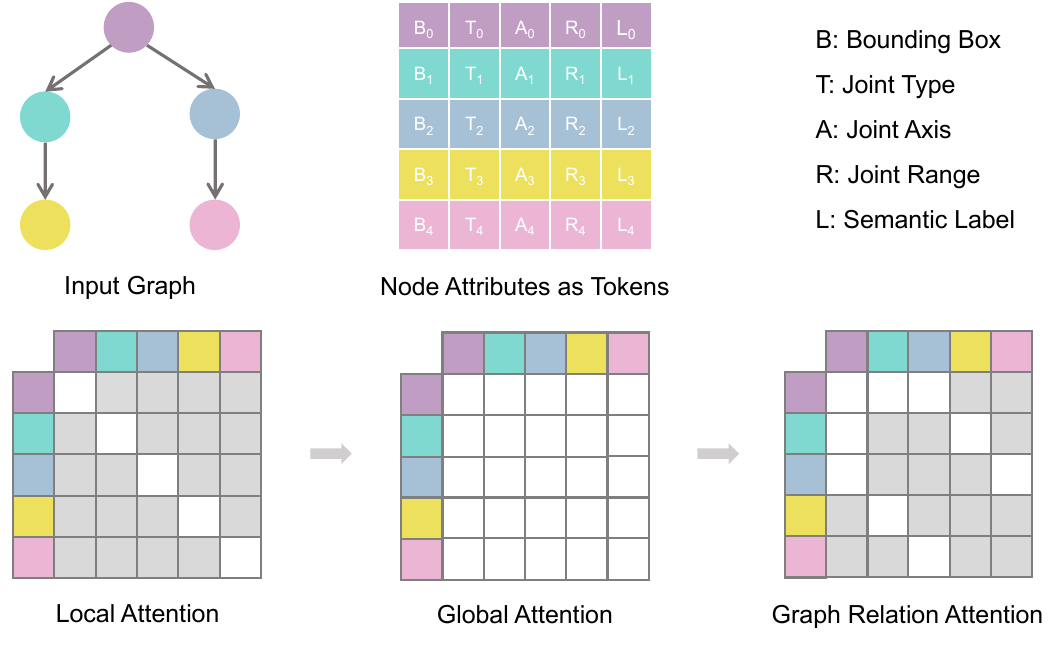}
    \caption{Design of the attention modules within our attribute attention blocks (AAB). Each node attribute is projected to a separate token and sequentially passed to three attention modules with varied masking strategies. White cells signify activated attention positions, whereas grey cells indicate attention that has been masked out.
    In graph relation attention, the activated cells represent the parent and child parts associated with each node.}
    \label{fig:attention}
\end{figure}

\mypara{Attention modules.}
Prior to being input into the attention layers, the attribute features $\hat{a}_{i,j}$, timestep features $\hat{t}$, and category features $\hat{c}$ are fused in the adaLN-Zero layer.
This is when the condition of timestep and object category gets injected.
The normalized and embedded tokens $\{\hat{f}^{t,c}_{i,j}\}$ are ready to go through three attention layers sequentially with structured masking (see \Cref{fig:attention}).
The intuition behind the design of each attention module is as follows.
\begin{itemize}\denselist
    \item \emph{Local Attention (LA)} captures the relationship between attributes within the part itself. 
    As the dependency between shape and articulation is both within and among parts, we intend to make node attributes carry the relationship among themselves to exchange information between nodes later. 
    The masking in LA only activates attention positions among attributes within the same node.
    \item \emph{Global Attention (GA)} allows for attention between every pair of attributes across all valid nodes. 
    The valid nodes are indicated by a key padding mask referring to the number of parts in each graph. 
    This module is designed to capture the relationship among nodes regardless of the distance in the graph. 
    Carrying the information from distant nodes to the subsequent module, which only concentrates on a local neighbourhood, can enhance overall hierarchical understanding.
    For masking, we only apply a key padding mask to ensure attention is applied only between non-padded attributes.
    \item \emph{Graph Relation Attention (GRA)} focuses on the relationship between attributes only from parent and children nodes.
    This is where we leverage the graph structure as a condition by explicitly masking the attention using the graph adjacency matrix.
    The root node (purple in \Cref{fig:attention}) is the sole self-connected node in the matrix.
    The model latches onto this signal through attention, and deduces edge direction by tracing down the tree from it.
    After propagating to the zero-ring and N-ring in the graph in the preceding attention layers, here we focus on one-ring relations, which are the strongest factor for articulations.
\end{itemize}
Our experiments show the effectiveness of these attention modules in \Cref{subsec:ablation}.

\begin{figure*}
    \centering
    \includegraphics[width=\textwidth]{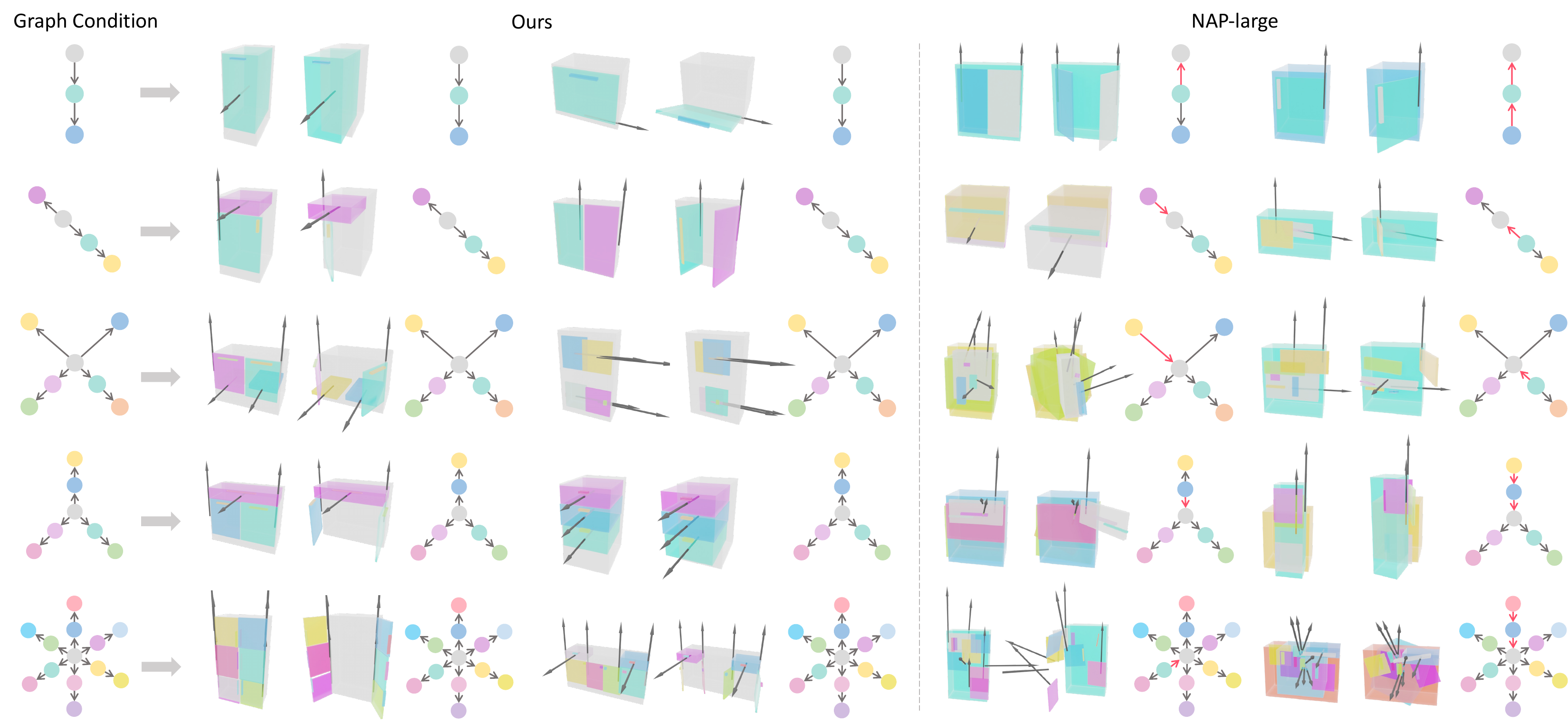}
    \vspace{-5mm}
    \caption{
    Qualitative results conditioned on graph structures (on the left) at different levels of complexity. 
    We compare our method with a comparable version of NAP. 
    Our generated objects are faithfully compatible with the graph input.
    In contrast, NAP fails to conform to the input constraint with flipped or disordered node connections.
    We denote inconsistent graph connections using \textcolor{brightred}{red} arrows.}
    \label{fig:cond_graph}
\end{figure*}

\section{Experiments}

\subsection{Experimental Setup}
We use the PartNet-Mobility dataset~\cite{mo2019partnet} with an 80/20 train-test split ratio per object category.
We use eight object categories (i.e., Storage, Table, Refrigerator, Dishwasher, Safe, Oven, Washer, and Microwave) with a maximum of $32$ nodes in the \textit{action} graph.
The architecture of the denoising network stacks $12$ layers of attribute attention blocks with $32$ heads in each attention module.
We train with the AdamW~\cite{loshchilov2018decoupled} optimizer ($\beta_1=0.9$, $\beta_2=0.999$, $\lambda=0.01$) for $5000$ epochs, taking $13$ hours training on a single NVIDIA A40 GPU with batch size as $64$.

\begin{table}
\centering
\resizebox{0.9\linewidth}{!}{
\begin{tabular}{@{}cccccc@{}}
\toprule
& \multicolumn{2}{c}{MMD} & \multicolumn{2}{c}{COV} & ~~Realism~~ \\ 
\cmidrule(l){2-3} \cmidrule(l){4-5} \cmidrule(l){6-6}
                  & ID$\downarrow$  & AID$\downarrow$& ID$\uparrow$     & AID$\uparrow$    & AOR$\downarrow$  \\ \midrule
NAP~\cite{lei2023nap}               & 0.118           & 0.751          & 0.752            & 0.794            & 0.026            \\
NAP-light         & 0.060           & 0.741          & \textbf{0.773}   & 0.866            & 0.062             \\
Ours-light        & \textbf{0.043}  & \textbf{0.636} & 0.753            & \textbf{0.867}   & \textbf{0.007}    \\ \bottomrule
\end{tabular}
}
\caption{
We evaluate the distribution modeling of a variation of our approach and compare against NAP and its variation when training on \textit{articulation} graph with $K=8$.
Our approach largely outperforms the baselines in terms of similarity to ground truth test set samples (measured by the MMD metrics), coverage of the ground truth test set distribution (measured by the COV metrics), and realism as measured by the AOR metric.
}
\label{tab:quant_k_8}
\end{table}

\begin{table}
\resizebox{\linewidth}{!}{
\begin{tabular}{@{}ccccccc@{}}
\toprule
& \multicolumn{2}{c}{MMD} & \multicolumn{2}{c}{COV} & \multicolumn{2}{c}{Realism} \\ 
\cmidrule(l){2-3} \cmidrule(l){4-5} \cmidrule(l){6-7}
                  & ID$\downarrow$  & AID$\downarrow$   & ID$\uparrow$    & AID$\uparrow$    & AOR $\downarrow$   & HS \% $\uparrow$  \\ \midrule
NAP-large         & 0.067           & 0.860             & 0.716           & 0.716            & 0.034              & 17.39      \\ 
Ours              & \textbf{0.049}  & \textbf{0.816}    & \textbf{0.753}  & \textbf{0.852}   & \textbf{0.008}     & \textbf{82.61}      \\ \bottomrule
\end{tabular}
}
\caption{
We evaluate the distribution modeling of our model and compare against NAP's variation when training on \textit{action} graph with $K=32$.
In this more challenging setting requiring handling of complex input graphs, our approach significantly outperforms the NAP baseline along all metric axes, including in perceived generation realism as measured by human judgment, reported in HS column as a preference rate.
}
\label{tab:quant_k_32}
\end{table}

\subsection{Baselines}

We compare our approach with NAP~\cite{lei2023nap} and several variants.
NAP's data representation differs from ours as: 
1) an additional 128$D$ latent code is used to represent the surface for each part; 
2) the joint type is not explicitly modelled.
The specific baseline variations are:
\begin{itemize}
    \item \textbf{NAP}: the original NAP model and data representation, trained on \textit{articulation} graphs ($K=8$).
    \item \textbf{NAP-light}: NAP with our data parameterization, trained on \textit{articulation} graphs ($K=8$).
    \item \textbf{Ours-light}: variant of our method trained on \textit{articulation} graphs with $K=8$ to be comparable with NAP-light.
    \item \textbf{NAP-large}: uses our data parameterization and is trained on \textit{action} graphs with $K=32$.
\end{itemize}

\subsection{Metrics}
\label{subsec:eval_metrics}
Our evaluation metrics rely on a notion of distance between articulated objects.
We use the Instantiation Distance (\textbf{ID}) from NAP~\cite{lei2023nap} which considers both part geometry and motion.
This is the minimum pairwise Chamfer-L1 distance per articulation state using 2048 point samples per part per object.

We refine this metric to consider the pairwise distance for temporally synchronized states, rather than enumerating across all state pairs as in \citet{lei2023nap}
This refinement enhances computation efficiency and eliminates spurious point pair distance computations caused by erroneous part matching.
We also introduce the Abstract Instantiation Distance (\textbf{AID}), which minimizes the influence of fine-grained part geometry by using volumetric IoU (vIoU) on part bounding boxes instead of Chamfer distance.

Armed with these two distances we define the following evaluation metrics:
1) Minimum Matching Distance (\textbf{MMD}) reports the average minimum matching distance between the ground truth set and the generated set. 
It measures the similarity between the approximated and ground truth distribution.
2) Coverage (\textbf{COV}) is the percentage of ground truth objects with at least one matched generated sample. 
A large value indicates better diversity of the generation and better distribution coverage.
3) 1-Nearest Neighbor Accuracy (\textbf{1-NNA}~\cite{lei2023nap}) reported using AID. This metric measures the distance between the generated and ground truth distribution using 1-NN classification accuracy.

We also report the Average Overlapping Ratio (\textbf{AOR}), computed as the average ratio of overlapping volume between any two sibling parts in the object.
We design this metric by assuming that the sibling nodes in the graph should never overlap in any articulation state.
Overlapping volumes are determined from part oriented bounding boxes and computed using vIoU.
This metric measures the physical plausibility of the generated part structure.
Lastly, we report human judgments of generated object quality and plausibility through a human study (\textbf{HS}).
We conducted this study using a two-alternative forced choice setup (A/B choice).
Participants were shown 20 pairs of randomly generated results from ours and NAP-large, and asked to choose the object with highest quality in terms of part articulations, part arrangement (i.e. no overlaps), and plausibility compared to real-world objects.
We collected responses from 44 participants not involved with this work and report the preference rate in \Cref{tab:quant_k_32}.

\subsection{Part Retrieval}

Once we have a generated articulated object abstraction specifying all the parts and their attributes (and optionally an object category constraint specified in the input), we use a part retrieval strategy to extract part surfaces from the training data and build the final 3D object mesh.
The training data contains a total 527 objects composed of 2690 parts which can be composited into a generated object through part-level retrieval.
We use a two-step approach:
1) we compute the AID metric of the generated object against candidate objects in the train set.
We identify the candidate with the best AID metric and pick the base part from it.
This procedure ensures that the base part is compatible with the part motions as much as possible.
2) For the remaining parts, we pick as many parts as possible from a single candidate object to maintain style consistency.
We start with the candidate selected in Step 1 and consider other candidates while there are still parts left unretrieved. 
Please refer to the supplement for more details.

\section{Results}

\subsection{Overall Generation Quality}
We evaluate how well our method models the real data distribution by randomly generating samples conditioned on the category labels and graphs in the testing set and computing the metrics described in \Cref{subsec:eval_metrics} against the test objects.
We generate five times as many objects as the test set and report the results in \Cref{tab:quant_k_8,tab:quant_k_32}. 
\Cref{tab:quant_k_8} compares baseline models trained on articulated parts with $K=8$.
\Cref{tab:quant_k_32} shows a more challenging setting that includes actionable parts and more complex graph topology, with $K=32$.
The lower MMD and higher COV values demonstrate our method better captures the training data distribution than NAP and other baselines.
Our method outperforms both NAP-large (0.567 vs. 0.728) and NAP-light (0.495 vs. 0.521) on 1-NNA as well. In addition,
our lower AOR indicates that generated objects suffer less from overlapping volumes among sibling parts during articulation, which suggests more physically realistic objects.
The HS score further supports the finding that our generated objects are of higher fidelity and realism, as perceived by people.

\begin{figure}
    \includegraphics[width=\linewidth]{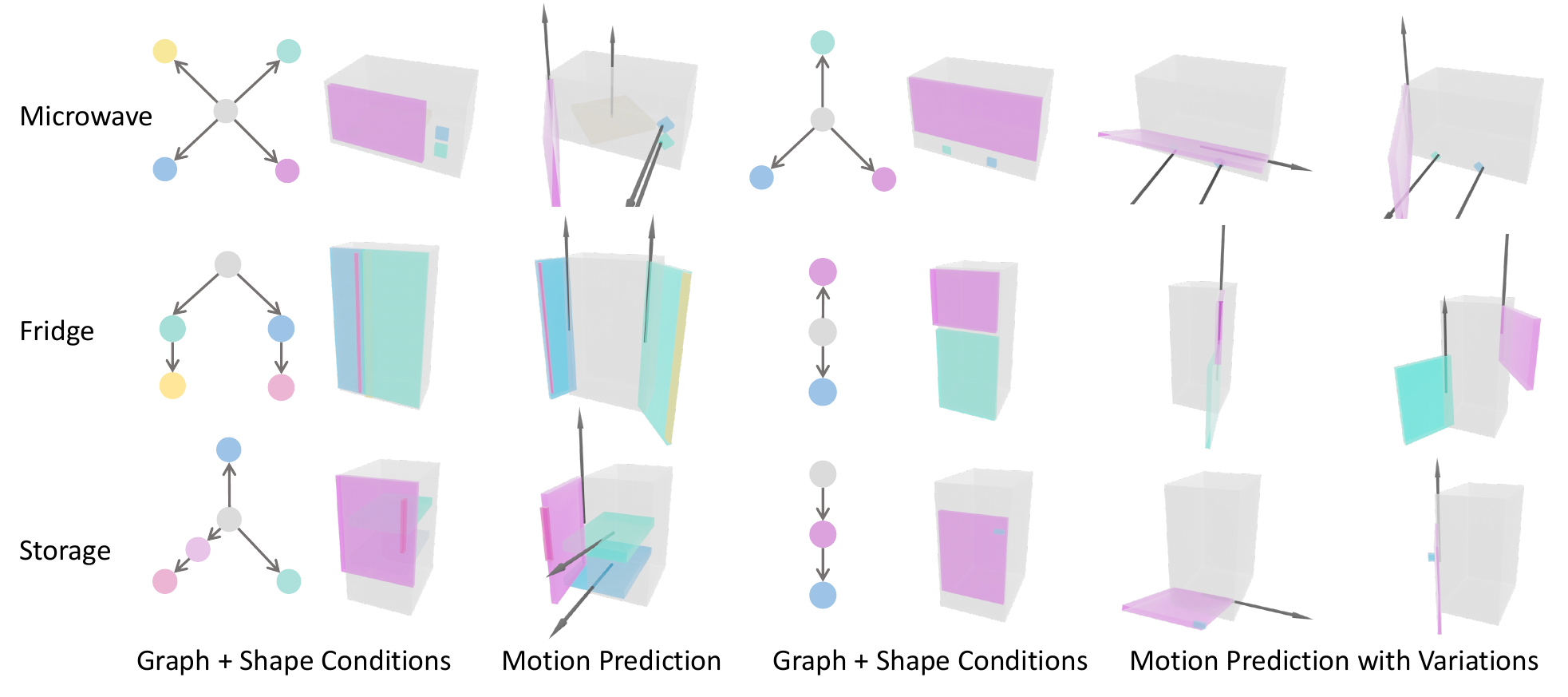}
    \caption{Part$\rightarrow$Motion: generated results conditioned on graphs specifying part bounding boxes.}
    \label{fig:cond_box}
\end{figure}

\begin{figure}
    \includegraphics[width=\linewidth]{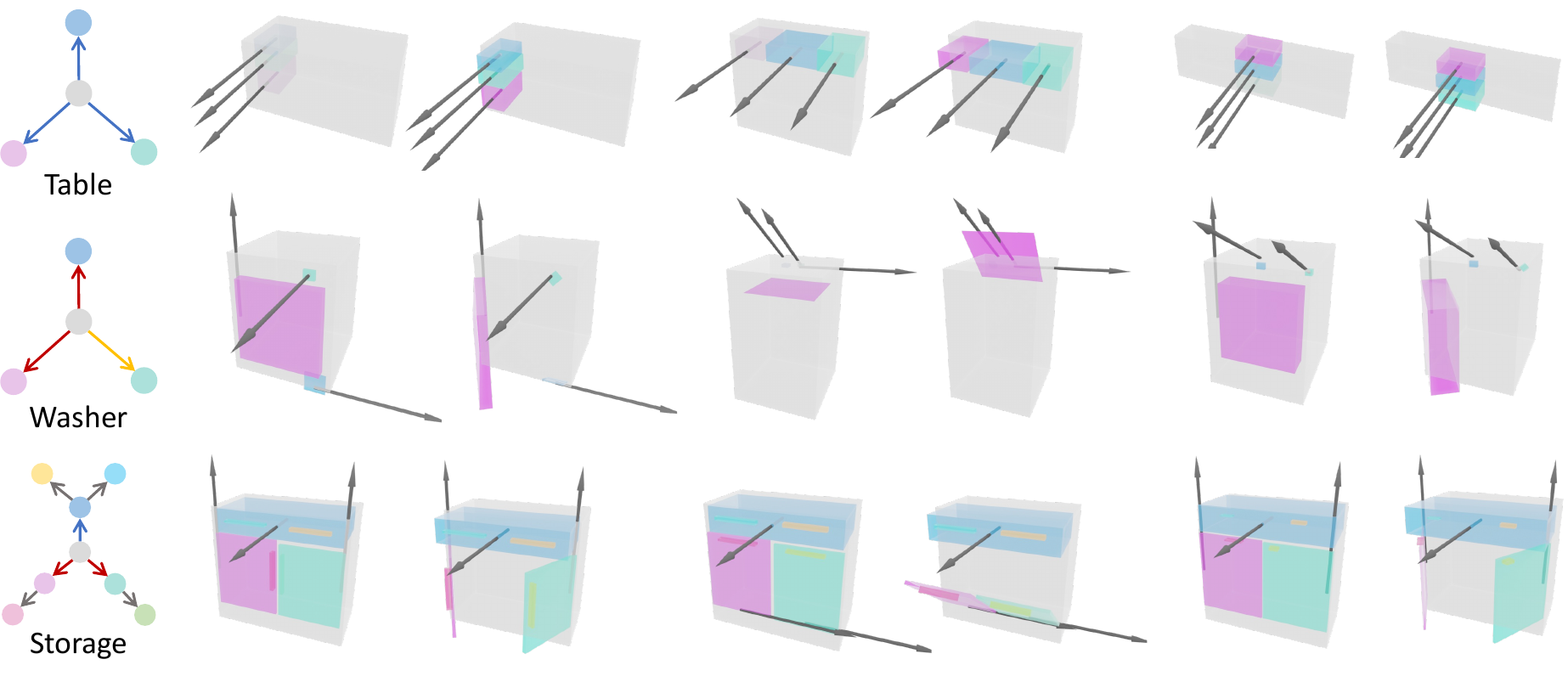}
    \vspace{-1em}
    \caption{Joint Type $\rightarrow$ Part: qualitative results conditioned on specific articulation joint type.
    Edge colors in the graph indicate the input joint type: \textcolor{jointblue}{blue} for prismatic, \textcolor{jointred}{red} for revolute, \textcolor{jointyellow}{yellow} for continuous, and \textcolor{jointgrey}{grey} for fixed. The outputs conform to these constraints while exhibiting variety.}
    \label{fig:cond_type}
\end{figure}

\begin{figure}
    \includegraphics[width=\linewidth]{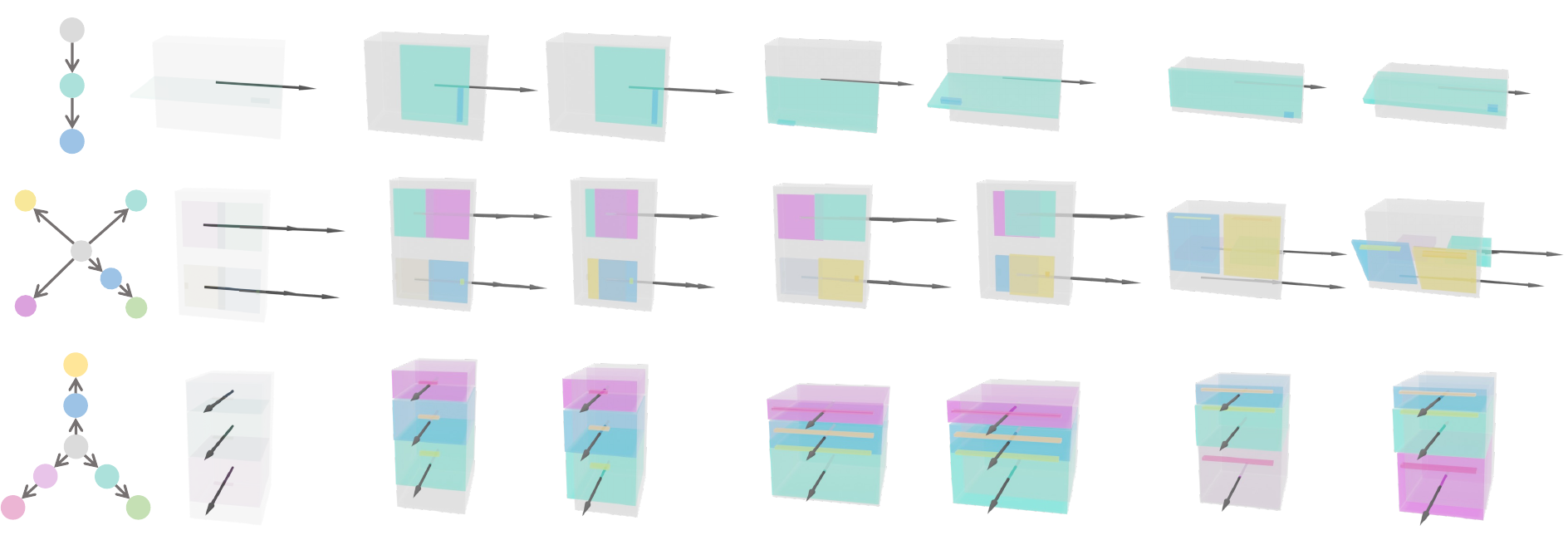}
    \caption{Joint Axis $\rightarrow$ Part: generated objects for input graphs specifying joint axis constraints (shown by arrows).
    Output objects have parts with motions corresponding to the given axes but varying part type and overall arrangement.}
    \label{fig:cond_axis}
\end{figure}

\begin{figure}
    \includegraphics[width=\linewidth]{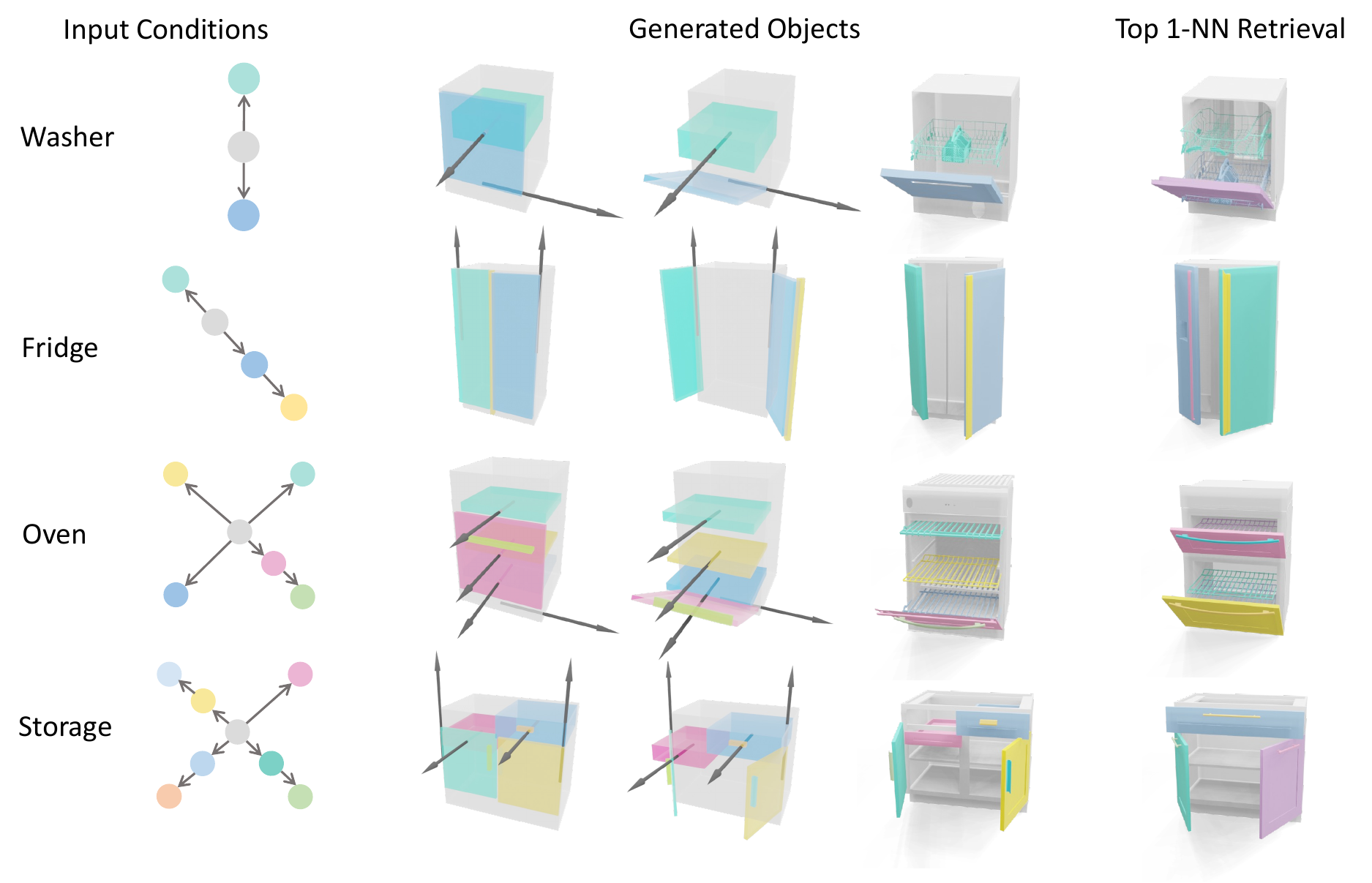}
    \vspace{-2em}
    \caption{Generated objects for out of distribution input graphs (structure unseen in train set) and the corresponding nearest neighbor retrieval from the train set. The generated objects for various categories exhibit realistic arrangement even though the exact number of parts and their arrangement were not in the train set.}
    \label{fig:ood}
\end{figure}

\subsection{Conditional Generation}

A key focus of our work is controllability.
Thus, we evaluate performance in five conditional generation tasks illustrating different forms of control over the output.
For graph-conditioned generation, we explicitly inject the graph structure into the denoiser as the condition.
For generation conditioned on other attributes associated with the graph, we mask the conditioning attributes with ground truth values and fill in other node attributes via denoising (akin to ``inpainting'').
For NAP, all the conditional generation are implemented using ``inpainting'' generation as above.
Note that conditioning on particular node attributes presupposes the inclusion of the graph as a condition.

\mypara{Graph-conditioned generation.}
\Cref{fig:cond_graph} shows qualitative examples at different levels of complexity for the input conditioning graph topology.
Our results are consistently plausible and of high quality, regardless of graph complexity.
In contrast, NAP often fails to respect the input graph topology and generates objects and graphs with flipped and disordered node connections.
NAP has a notably high rate of generating inconsistent graphs with the input condition, and this issue becomes even more pronounced when the graph constraints are more complex.

\mypara{Part $\rightarrow$ Motion.}
Given the position and size of the bounding box for each part, this experiment aims to generate compatible motion parameters (joint type, joint axis, and joint range).
\Cref{fig:cond_box} shows the qualitative results under different categories.
On the left, we present three examples that are expected to generate more deterministic motions, given the specified arrangement of parts.
On the right, we show generated motion with variations, each reasonably compatible with the specified shape conditions. 

\mypara{Joint Type $\rightarrow$ Part.}
Given the joint type for each part, this experiment aims to generate compatible bounding boxes and joint parameters (joint axis and joint range).
We show qualitative results in \Cref{fig:cond_type}.
Given the specified joint type (denoted in varied colours on the graph edges) for each part, we show variations of generated shapes adhering to these conditions.

\mypara{Joint Axis $\rightarrow$ Part.}
In this scenario, the position and direction of the joint axis for each part is given and we generate compatible bounding boxes, joint types, and joint ranges.
\Cref{fig:cond_axis} shows some results, demonstrating variation while conforming to the axis constraint.
Note that in cases where a part is connected via a prismatic joint, the part considers only the direction of the axis as its constraint.

\mypara{OOD graph-conditioned generation.}
This experiment shows how well our model generalizes the graph constraint by conditioning the generation on graph topologies that are out of the distribution of the training samples. 
\Cref{fig:ood} shows the qualitative results for various object categories.
These results illustrate that our method can generate reasonable objects even for unseen graphs as conditions.

\begin{figure}
    \includegraphics[width=\linewidth]{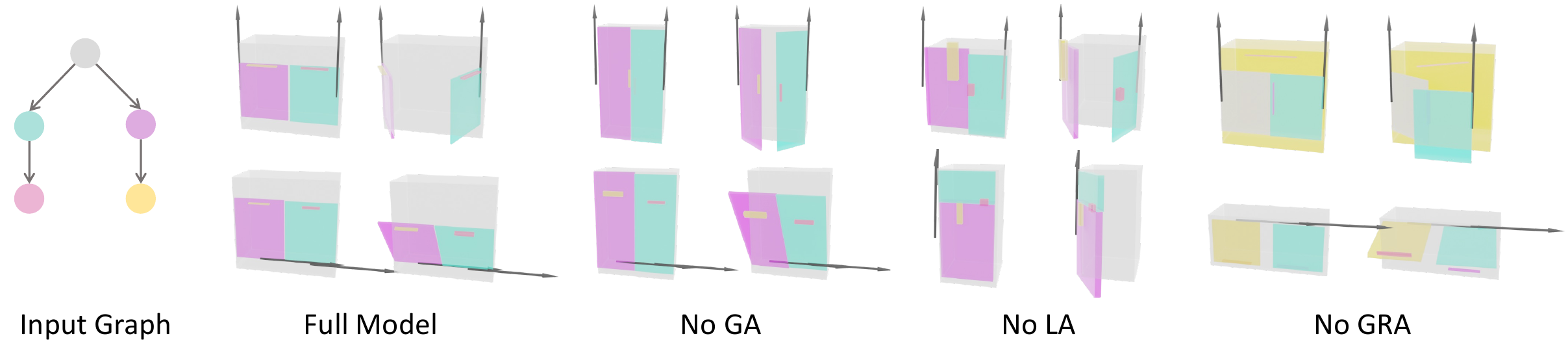}
    \caption{Ablation of each attention module in our denoising architecture.
    The significance of each module increases from left to right.
    Removing these modules leads to lower quality objects, with inconsistent connections between articulating parts, misassigned object base part (i.e. non-moving root for the object), and unrealistically floating part motions.
    }
    \label{fig:ablation}
\end{figure}

\subsection{Ablations}
\label{subsec:ablation}

\begin{table}
\centering
\begin{tabular}{@{}cccccc@{}}
\toprule
                    & full                & no LA     & no GA           & no GRA        \\ \midrule
MMD-AID$\downarrow$ & \textbf{0.816}      & 0.840     & 0.831           & 0.876          \\
MMD-ID$\downarrow$  & \textbf{0.049}      & 0.053     & 0.052           & 0.157          \\ \midrule
AOR$\downarrow$     & \textbf{0.008}      & 0.016     & 0.011           & 0.013          \\ 
\bottomrule
\end{tabular}
\caption{
Quantitative results for ablations of our attention modules.
MMD increases as individual attention modules are removed, indicating their positive impact on the design of our architecture.
There is also a corresponding drop in sample realism, as indicated by the increasing AOR metric.
}
\label{tab:ablation}
\vspace{-3mm}
\end{table}

To show the effectiveness of our architecture we ablate the local attention (LA), global attention (GA), and graph relation attention (GRA) modules in each attribute attention block (AAB) (see \Cref{fig:ablation,tab:ablation}).
\Cref{fig:ablation} shows qualitative examples conditioned on the same graph across ablations removing one module at a time.
The significance of each module increases progressively from left to right.
GA is designed to learn the relation between nodes beyond the 1-ring neighborhood (pink and yellow nodes in this case).
Removing GA makes handles less symmetric compared to the full model.
LA is designed to learn the relation between attributes within each part (e.g., handle should be at far end from joint axis).
By removing LA, this correlation is weakened.
GRA is the main module for learning part arrangements that conform to the input graph topology.
By removing GRA, the generated parts do not respect the specified part hierarchy.
In \Cref{tab:ablation}, the MMD score increases as modules are removed, indicating lower quality objects.
There is also a corresponding drop in sample realism, as indicated by the AOR score.
Generally, the removal of explicit attention to part relations leads to generated objects that tend to not conform to input constraints, making coverage-based metrics less meaningful.


\begin{figure}
\includegraphics[width=\linewidth]{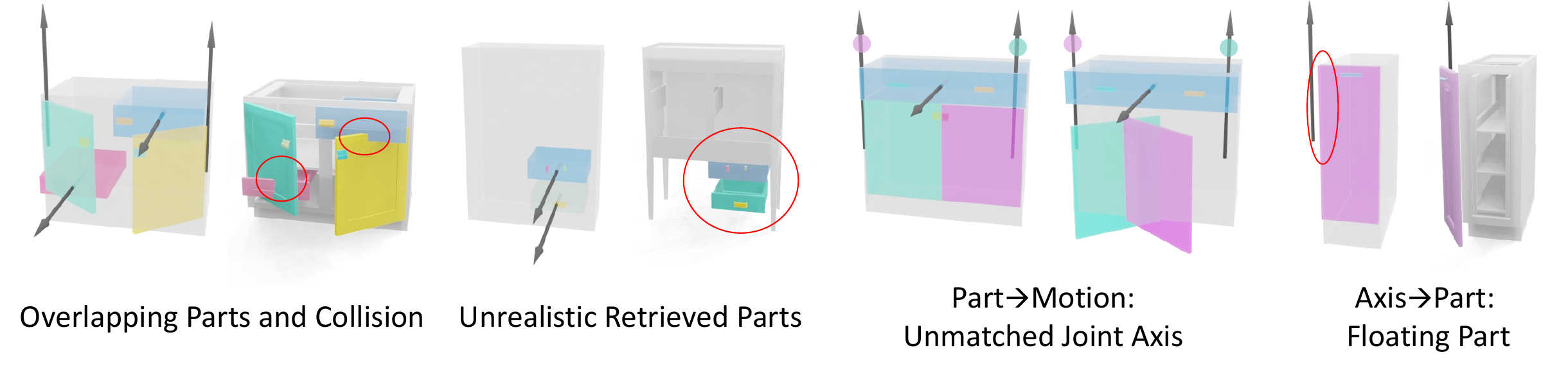}
\caption{We show several failure cases in terms of both generation quality and controllability.}
\label{fig:fail}
\end{figure}

\subsection{Failure Cases and Limitations}
\Cref{fig:fail} shows typical failure patterns in terms of generation quality and controllability.
1) Overlapping parts, or collisions during motion.
2) Reliance on part retrieval can lead to unrealistic part combinations (e.g., inconsistent drawers under desk).
3) Mismatched parts and joint axes in the \emph{Part $\rightarrow$ Motion} conditional scenario, leading to physically unrealistic motion.
4) Floating parts in the more challenging \emph{Joint Axis $\rightarrow$ Part} setting due to generated part not perfectly aligning with joint axis.

\section{Conclusion}
\label{sec:conclusion}

We address conditional generation of articulated 3D objects, allowing for fine-grained user-specified constraints on object parts and articulation.
We develop a denoising diffusion architecture with a set of part attribute attention blocks that guide generation based on the input conditions.
We thus leverage relations between part attributes and generate higher-quality objects that better conform to the user constraints compared to prior work.
Our qualitative and quantitative evaluations show that we significantly outperform the state-of-the-art, generating more realistic and more complex articulated objects, and exhibiting greater diversity.

Some limitations of our work suggest future work directions.
Due to significant data imbalance, our method has better performance on higher frequency objects.
Data augmentation schemes for obtaining a more uniform performance would be an interesting direction to explore.
Similarly to NAP~\cite{lei2023nap}, we face the challenge that motion attributes exhibit relatively weaker controllability than geometric attributes. While we enhance the dependency through attention among attributes, further investigation into reinforcing this connection is warranted.
Additionally, our part synthesis currently relies on a retrieval strategy which is limited by the diversity of available objects and parts.
Combining our work with part geometry generation is another interesting avenue for future work.

We believe that the fine-grained controllable generation of 3D articulated objects that our approach provides will enable scalable generation of interactive 3D assets in support of tasks in computer vision, robotics, and embodied AI.

\mypara{Acknowledgements.}
This work was funded in part by a Canada Research Chair, NSERC Discovery, and enabled by support from \href{https://www.westgrid.ca/}{WestGrid} and the \href{https://alliancecan.ca/}{Digital Research Alliance of Canada}. We thank Xiang Xu, Yizhi Wang, and Hanhung Lee for helpful discussions, as well as Angel X. Chang, Sonia Raychaudhuri, Xiaohao Sun, Qirui Wu, Yiming Zhang, and Ning Wang for proofreading.

{
\small
\bibliographystyle{ieeenat_fullname}
\bibliography{main}
}

\clearpage
\setcounter{page}{1}
\maketitlesupplementary

In this supplement to the main paper, we provide implementation details (\Cref{sec:appendixA}) and additional qualitative and quantitative results (\Cref{sec:appendixB}).
We also include a supplemental video to provide a quick introduction to our work.

\section{Implementation Details}
\label{sec:appendixA}

\subsection{CAGE Training Details}
We randomly permute the ordering of the nodes in graphs during training so that our model does not learn to rely on specific node orderings. 
This makes our model robust to graph isomorphic permutations at inference time.
For parameterization, we normalize all the part attributes in the range of $[-1, 1]$ to feed into the diffusion model.

Our diffusion model follows the standard scheme proposed in DDPM~\cite{ho2020ddpm}. 
In the forward process, we use a linear beta scheduler that maps a sequence of betas ranging from $1e-4$ to $0.02$ with a total of $1,000$ diffusion steps.
Each training iteration consists of $64$ objects and each object is trained with $10$ randomly sampled timesteps.
We set an initial learning rate of $5e-4$.
To schedule the learning rate, we use a warm-up strategy for $20$ epochs and then decay from the initial learning rate to $0$ by following the cosine function.
During inference, we set $100$ denoising steps for generating each sample.
We train the diffusion model for $5,000$ epochs in total on a single NVIDIA A40 GPU for $13.94$ hours.
We show a quantitative comparison of the training time with other baselines in \Cref{tab:train_time}.

\begin{table}[h]
\centering
\resizebox{\linewidth}{!}{
\begin{tabular}{@{}cccccc@{}}
\toprule
& \multicolumn{3}{c}{$K=8$, \textit{articulation} graph} & \multicolumn{2}{c}{$K=32$, \textit{action} graph}  \\ 
\cmidrule(l){2-4} \cmidrule(l){5-6}
                        & NAP~\cite{lei2023nap} & NAP-light      & Ours-light      & NAP-large      & Ours    \\ \midrule
Training time (hrs)     & 6.05                  & 1.76           & 2.74            & 22.87          & 13.94   \\
\bottomrule
\end{tabular}
}
\caption{
Training time comparison with baselines and variations under two experiment settings using 32-bit float precision.
\textbf{Setting 1}:
for models with $K=8$, the \textit{shapecode} formulation in the NAP model accounts for the majority of the computational overhead, as evidenced by the comparison with the NAP-light variant which replaces this component with a lighter part type attribute.
The comparable variant of method (Ours-light) incurs a marginally higher computation time than NAP-light, while providing significantly better outputs as described in the main paper.
\textbf{Setting 2}:
in scenarios where the methods are extended to more parts ($K=32$), the computational overhead for the NAP-large model escalates significantly whereas our method's training time remains comparatively lower.
These results demonstrate the improved computational scalability of our approach relative to NAP, which is important for complex objects with larger numbers of distinct parts.
}
\label{tab:train_time}
\end{table}

\subsection{Part Retrieval Implementation}

Given a generated articulated object abstract specification, we use a two-step approach to retrieve suitable parts and build the final articulated 3D object.
We pick the base part in step 1 and the remaining parts in step 2.

\mypara{Step 1: base part retrieval.}
We compute a Weisfeiler-Lehman graph hash~\cite{shervashidze2011weisfeiler} from the generated object kinematic tree.
This hash is identical for isomorphic graphs.
Since the base part (i.e. stationary part of the object) should be compatible with the generated object part motions, we anticipate that the best--matching object candidates will have the same kinematic tree topology.
Hence, only candidates with the same hash (and object category, if specified in the input) from the training set are selected for further consideration.
In cases where no candidates have the same hash (which happens when the object is out of distribution), we consider all candidates in the training set.
We then compute the AID metric for each selected candidate and pick the base part from the candidate with best metric value.
In addition, we keep the top five candidates for the next step.

\mypara{Step 2: articulated part retrieval.}
For the remaining parts other than the base, we pick a single candidate part for each semantic part in the generated object (e.g., one drawer part for a storage with three drawers).
The selected part is duplicated and resized to fill the part bounding box in the generated object abstract specification.
We start with parts in the top five object candidates in step 1 and choose a part if it has the same semantic label as any of the required parts.
If there are still unretrieved parts, we consider other candidates from the same category (if specified) or the whole training set.

The above procedure is designed to maximize style consistency between the retrieved parts, and create coherent objects.
The full 3D mesh visualizations in the main paper and in this supplement demonstrate the results obtained using this approach.

\subsection{Metrics}

Here we provide implementation details for the two distances used in our evaluation metrics.

\mypara{Instantiation Distance (ID).}
We simplify the ID metric first proposed by NAP~\cite{lei2023nap} to consider pairwise Chamfer-L1 distance between two objects in temporally synchronized articulation states.
Specifically, we randomly sample 2,048 point samples per part per object and compute their Chamfer distance in five evenly spaced-out articulation states within the joint ranges of the objects.
We take the average of the five distances to be the final ID value.

\mypara{Abstract Instantiation Distance (AID).}
We also introduce AID to measure the distance between two objects with volumetric IoU (vIoU) on the part bounding boxes.
Given two objects, we first scale both objects such that their overall bounding boxes are the same size.
We then assign part correspondences between the two objects based on the part bounding box center distances.
For each part pair, we compute a sampling-based vIoU (using $10,000$ point samples per bounding box) in five evenly spaced-out articulation states within the joint ranges of the objects.
Finally, we take the average of the vIoUs over all states and parts and compute the complement ($1 - \text{vIoU}_\text{avg}$) as the final AID value.
\section{Additional Results}
\label{sec:appendixB}

\begin{figure*}
    \includegraphics[width=\textwidth]{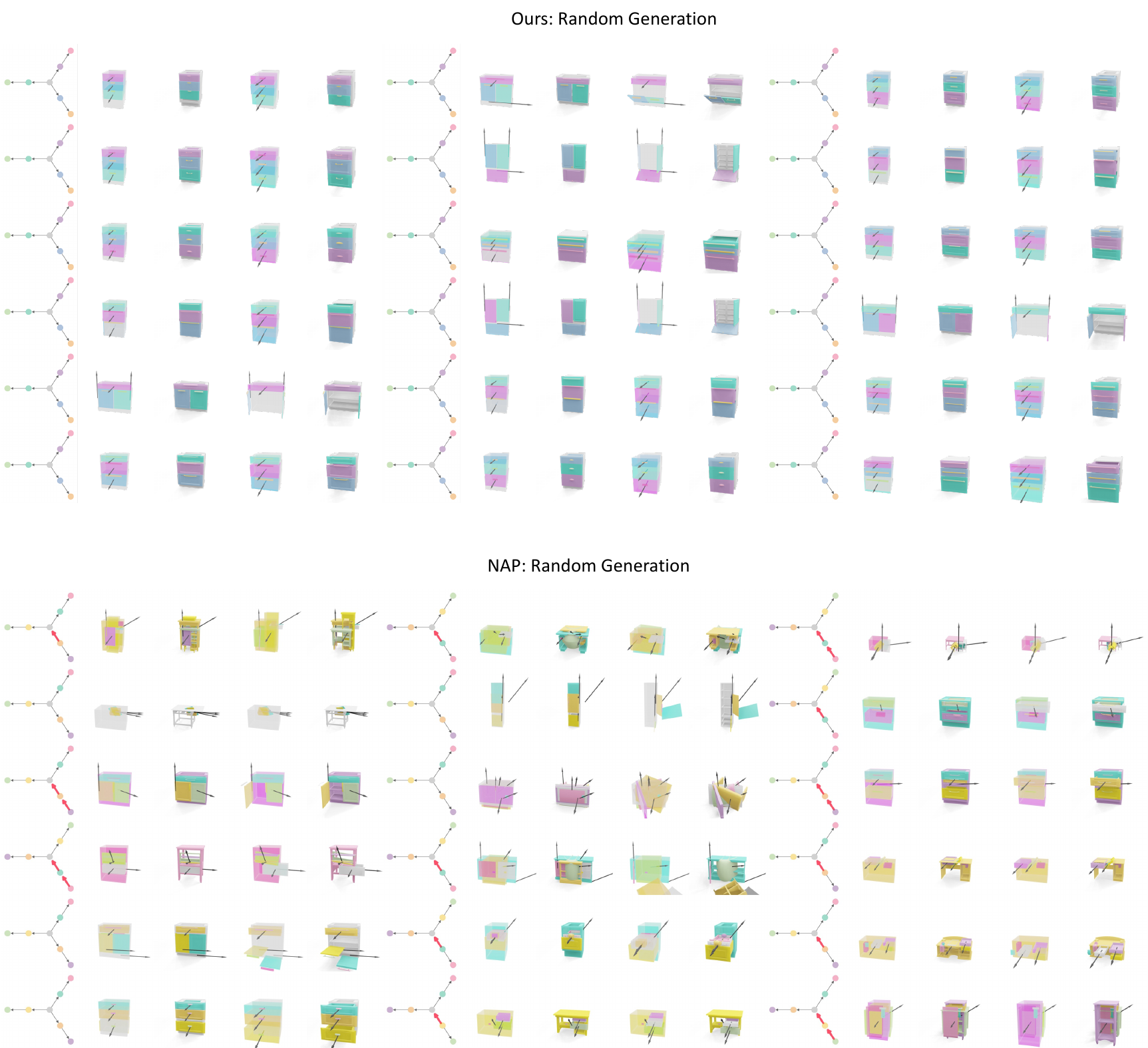}
    \caption{
    \textbf{Graph conditional generation}:
    we show 18 randomly generated samples (without manual selection) produced with our method (top) and NAP-large (bottom).
    \textbf{Summary}:
    Our results are consistently compatible with the input graph and achieve overall high fidelity.
    In comparison, the results from NAP often fail to conform to the graph constraint with many flipped edge connections denoted by \textcolor{brightred}{red} arrows in the output part hierarchy graph.
    Moreover, many objects generated with NAP-large exhibit unrealistic part-part overlaps and incorrectly oriented motion axes.
    \textbf{Setup}:
    The results from ours and NAP-large are both conditioned on a medium-complexity object articulation graph identical to the graph shown in the top-left entry.
    The results are selected from the first 18 results generated from each model.
    For every set of five columns, the first column shows the node hierarchy of the generated parts.
    The second and third columns depict the object in its resting state in abstract and complete mesh form.
    Columns four and five represent the object in a fully open state, again in abstract and mesh form.
    }
    \label{fig:supp:rand_cond_graph}
\end{figure*}

\mypara{Graph conditional generation.}
\Cref{fig:supp:rand_cond_graph} shows 18 randomly generated samples (with no manual selection) from our method CAGE, and the baseline NAP-large adapted from \citet{lei2023nap}.
Both are conditioned on medium-complexity object graphs.
The results are the first 18 results generated from each method.
The output objects generated using our method are consistently compatible with the input graph and achieve overall high fidelity.
In comparison, the objects generated using NAP often fail to conform to the graph constraint with flipped edge connections, denoted in red arrows.
Moreover, many of the objects generated using NAP are unrealistic, with incorrect articulation axes and significant part-part misorientations and collisions.

\begin{figure*}
    \includegraphics[width=\textwidth]{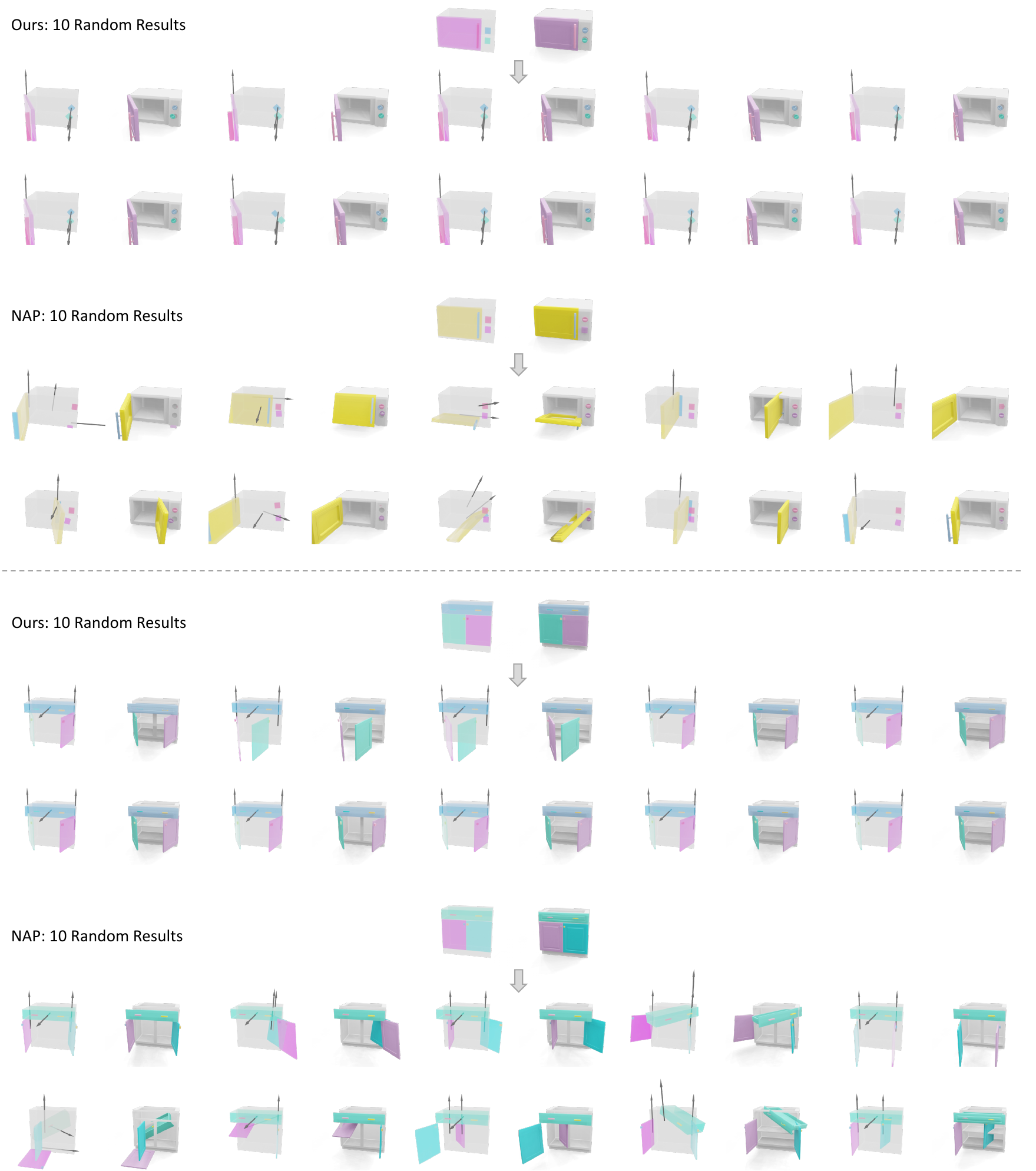}
    \caption{
    \textbf{Part $\rightarrow$ Motion}:
    randomly generated samples (with no manual picking) from our method and NAP-large for two example inputs shown at the top.
    \textbf{Summary}:
    our generated motions are stable and consistently compatible with the input part shape (e.g. in the microwave example, the handle on the right of the door indicates that the door is more likely to be opened from the right).
    The second example is more challenging with more parts to be coordinated.
    As expected, we observe some failure cases in our results, with implausible articulation motions (see outputs in the sets at column 3 and column 5, from the left).
    In this more challenging case, NAP by comparison produces implausible motions in a much larger proportion of the generated results with almost all generated objects exhibiting part collisions or unrealistic motion axes.
    \textbf{Setup}:
    Every set of two columns shows a generated object, with the first column showing the abstract bounding box form with joint axes and the second column showing the final retrieved part meshes, both in the fully open state.
    }
    \label{fig:supp:rand_cond_box}
\end{figure*}

\mypara{Part$\rightarrow$Motion.}
\Cref{fig:supp:rand_cond_box} shows two sets of randomly generated samples (with no manual selection) from our method and NAP-large that are conditioned on part shape attributes.
The generated motion using our method is stable and consistently compatible with the input part shape (e.g. in the microwave example, the handle on the right of the door indicates that the door is more likely to be opened from the right).
The second example is a more challenging case with more parts to be coordinated.
Here, we observe a few failure cases with implausible articulations in our results (columns 3 and 5 from the left).
In comparison, a large proportion of the results from NAP exhibit implausible motions.



\end{document}